\documentclass[conference]{IEEEtran}
\IEEEoverridecommandlockouts
\usepackage{cite}
\usepackage{amsmath,amssymb,amsfonts}
\usepackage{graphicx}
\usepackage{amsmath}
\usepackage{amssymb}
\usepackage{amsthm}
\usepackage{bm}
\usepackage{booktabs}
\usepackage{multirow}
\usepackage{algorithm,algpseudocode}
\usepackage{textcomp}
\usepackage{xcolor}
\usepackage{subcaption}
\usepackage{setspace}
\def\BibTeX{{\rm B\kern-.05em{\sc i\kern-.025em b}\kern-.08em
    T\kern-.1667em\lower.7ex\hbox{E}\kern-.125emX}}
\begin{document}

\title{A Hybrid Model for Learning Embeddings \\ and Logical Rules Simultaneously\\ from Knowledge Graphs}

 \author{\IEEEauthorblockN{Susheel Suresh and Jennifer Neville}
 \IEEEauthorblockA{Computer Science Department \\
 Purdue University\\
 West Lafayette, IN, USA \\
{[}suresh43, neville{]}@purdue.edu}
 %\and
% \and
% \and
% \and
% \and
}
\maketitle

\begin{abstract}
    The problem of knowledge graph (KG) reasoning has been widely explored by traditional rule-based systems and more recently by knowledge graph embedding methods. While logical rules can capture deterministic behavior in a KG they are brittle and mining ones that infer facts beyond the known KG is challenging. Probabilistic embedding methods are effective in capturing global soft statistical tendencies and reasoning with them is computationally efficient. While embedding representations learned from rich training data are expressive, incompleteness and sparsity in real-world KGs can impact their effectiveness. We aim to leverage the complementary properties of both methods to develop a hybrid model that learns both high-quality rules and embeddings simultaneously. Our method uses a cross feedback paradigm wherein, an embedding model is used to guide the search of a rule mining system to mine rules and infer new facts. These new facts are sampled and further used to refine the embedding model. Experiments on multiple benchmark datasets show the effectiveness of our method over other competitive standalone and hybrid baselines. We also show its efficacy in a sparse KG setting and finally explore the connection with negative sampling.

\end{abstract}

% \begin{IEEEkeywords}
% \end{IEEEkeywords}

\section{Introduction}
    Knowledge graphs (KGs) are large directed graphs where nodes represent concrete or abstract entities and edges symbolize relations for a pair of nodes. Many KGs containing millions of entities and relation types exist today viz. Freebase \cite{freebase}, YAGO \cite{suchanek2007yago}, Wikidata \cite{wikidata}, and Google Knowledge Graph \cite{googlekg} which are pivotal in reasoning about multi-relational data from different domains. One important reasoning problem is that of predicting missing relationships between entities (link prediction). Reasoning over KGs is particularly challenging in part due of their characteristic properties: large size, incompleteness, sparsity and noisy facts. Latent feature (a.k.a embedding) models learned probabilistically (RESCAL \cite{rescal}, TransE \cite{transe}, ComplEx \cite{complex} and RotatE \cite{rotate}) and inductive logic programming (ILP) inspired techniques which mine interpretable logical rules (WARMER \cite{warmer},  AIME+ \cite{galarraga2015fast}) are two prominent KG reasoning approaches. 

Relations in knowledge graphs adhere to certain constraints which enforce syntactic validity and typically follow deterministic connectivity patterns like equivalence, symmetry, inversion and composition. Rules which capture such patterns are precise, interpretable and can generalize well. 
% The below rule explains the above example and it is easy to see it's generalizing power for a new sport and player.
% {\small \[playsSport(V_1, V_2) \leftarrow  playsFor(V_1, V_3) \; \land \; clubType(V_3, V_2)\]}%
Drawbacks includes potential low coverage, mining inefficiency (large search space) and difficulty in mining quality rules from incomplete KGs.
% KGs also conform to various "soft" statistical patterns that have considerable predictive capacity. These kind of ``soft" global patterns are non-deterministic and using logical rules to express them would be futile. Embedding based models aim to learn such "soft" features automatically from data and are shown experimentally to perform well. 
Embedding methods aim to learn useful representations of entities and relations by projecting known triplets into low-dimensional vector spaces and maximizing the total plausibility of known facts in the KG. Embedding models are able to capture unobservable but intrinsic and semantic properties of entities and relations \cite{jenatton2012latent}. As reasoning with embeddings boils down to vector space calculations it is computationally efficient, but it can be inaccurate when the entities and relations are sparse or noisy. \cite{pujara2017sparsity}. 
% Because negative instances are not marked in a KG, \textit{negative sampling} is employed for learning embeddings with gradient descent. The way negatives are sampled have a strong influence on embedding learning \cite{kotnis2017analysis} because, it relates to data quality.

Since both methods have their advantages and disadvantages, in this paper we propose a hybrid model which aims to exploit their complementary strengths. The main idea is to selectively utilize inferred facts from logical rules for embedding learning. Also, embedding feedback is used to prune the search space of the rule mining system. This cross feedback process when run for many iterations, simultaneously learns to incorporate deterministic structure into embeddings and mine rules that are consistent with ``global" KG patterns.
Hybrid models proposed previously in the literature use simple logical rules to place constraints on the embedding space in order to incorporate structure. Rule learning is detached from embedding learning and most methods just naively enumerate all possible rules to start with, which doesn't scale for larger KGs. Methods like IterE \cite{zhang2019iterE} place assumptions on the kind of rules and embedding techniques that can be used and are built specifically to tackle sparse entities. Different from such methods, our model aims to simultaneously improve embeddings and mine diverse and reliable rules. Moreover, our method can be incorporated with any embedding technique and rule mining system. 

\section{Background}
    Let $\mathcal{E}$ represent the set of all entities and $\mathcal{R}$ the set of all relation types in the KG. A knowledge graph $G$ contains a set of factual triplets $\{(h, r, t) | h, t \in \mathcal{E}; r \in \mathcal{R}\}$. $h,r,t$ are called head, relation and tail respectively. Figure \ref{fig:sample_kg} shows a sample from a larger knowledge graph about sports. Knowledge graph reasoning in particular link prediction deals with the problem of inferring new relationships between entities and triplet classification involves predicting the existence of candidate triplets in a given KG.

\noindent \textbf{Knowledge graph embedding} 
Entities are represented as vectors and relations are seen as operations in vector space and are typically represented as vectors, matrices or tensors. These models assume that the existence of individual triplets in a KG are conditionally independent given latent representations a.k.a embeddings of entities and relations in a continuous vector space.
A score function $\phi : \mathcal{E} \times \mathcal{R} \times \mathcal{E} \rightarrow \mathbb{R}$ is used to measure the model's confidence in a candidate triplet and defined based on different vector space assumptions for e.g., a popular model called TransE \cite{transe} aims is to have $\mathbf{h} + \mathbf{r} \approx \mathbf{t}$ if $(h,r,t) \in KG$, (boldface letters represent respective embeddings in $\mathbb{R}^d$). So, the score function  $\phi(h,r,t) = -\Vert \mathbf{h} + \mathbf{r} - \mathbf{t} \Vert$ 
is expected to be large if $(h,r,t)$ exists in the KG . 
% Different scoring functions are shown in Table \ref{tab:score_functions}.
% RESCAL \cite{rescal} one of the earliest works in knowledge graph embedding, used a bilinear form for the scoring function inspired from matrix factorization. Specifically, $\phi(h,r,t) = \mathbf{h}^{\top} \mathbf{M_r} \mathbf{t}$  where entities are represented by vectors in $\mathbb{R}^d$ and relations as matrices in $\mathbb{R}^{d \times d}$ that model pairwise interactions.
Embedding learning is done under open world assumption (OWA) where un-observed triplets are either false or unknown. Negative sampling is employed due to the lack of negative examples in the input KG. A simple and effective approach \cite{transe} is to corrupt either the head or tail of a true triplet with a random entity sampled uniformly from $\mathcal{E}$. Entity and relation embeddings are learned by minimizing logistic or pairwise ranking loss.

\begin{table}
\small
\renewcommand{\arraystretch}{1.0} % Default value: 1
\caption{State-of-the-art embedding score functions.}
\centering
\begin{tabular}{lc}  
\toprule
Model  & Score Function ($\phi$)\\
\midrule
TransE & $-\Vert \mathbf{h} + \mathbf{r} - \mathbf{t} \Vert$\\  
DistMult & $\langle \mathbf{h},\mathbf{r},\mathbf{t} \rangle$\\
ComplEx & $\textbf{Re}(\langle \mathbf{h},\mathbf{r},\overline{\mathbf{t}} \rangle)$\\
RotatE & $-\Vert \mathbf{h} \circ \mathbf{r} - \mathbf{t} \Vert$\\
\bottomrule
\end{tabular}
\vspace{-4mm}
\label{tab:score_functions}
\end{table}

\noindent\textbf{Rule Mining}
% A First order logic (FOL) rule is a formulae of atoms connected with logical connectives $(\land, \lor, \lnot)$ and quantifiers $(\exists , \forall)$.
% An \textit{atom} such as $worksAt(X, Y)$ is a fact template that is made up of symbolic variables and/or constants. \textit{Grounding} of an atom is the instantiation of all its variables constants from $\mathcal{E}$, for eg. $worksAt(Bob, Google)$. 
% An example rule is:
% {
% \small
% \begin{align*}
% playsSport(V_1, V_2) \leftarrow playsFor(V_1, V_3) \; \land \; clubType(V_3, V_2) 
% \end{align*}
% }%
A rule is a formulae of \textit{atoms} connected with logical connectives. In particular, a rule is \textit{Horn} if the conjunction of a set of \textit{body} atoms results in a single head atom like, $\tau : B_1 \land B_2 \land \dots \land B_n \rightarrow r(X, Y) $ where $B_i$'s are body atoms and $r(X,Y)$ is a head atom. We use $body(\tau)$ to denote the body atoms of rule $\tau$. An \textit{instantiation} of a rule is the act of substituting all its variables with entities from $\mathcal{E}$. The head atom of an instantiated rule is called an \textit{inferred} head atom if all body atoms of the instantiated rule exist in the KG. $S_\tau$ is the set of all inferred head atoms obtained from the instantiation of a rule $\tau$.

\vspace{-3mm}
\begin{equation}
\label{eq:s_t}
    S_\tau = \{r(X,Y) \mid \exists z_1,\dots, z_m : body(\tau)\}
\end{equation}
\vspace{-4mm}

\noindent where $z_1,\dots, z_m $ are variables that appear in the body of the rule. A principled approach to mine horn rules from a KG is by an association rule learning algorithm \cite{galarraga2013amie} and \cite{warmer}. To deal with the vast search space, language biases viz. limiting rule length, having every atom to be transitively connected to each other in a rule (a.k.a \textit{connected} rules) and  ensuring all variables in a rule to to be \textit{closed} (i.e. appears at least twice) are utilized. During mining, various statistical measures assess the quality of intermediate rules to help prune the search space. According to \cite{galarraga2013amie}, 
\textit{Rule Support} is the number of distinct grounding of the head atom resulting from the body.  Formally, 
\begin{equation}
    supp(\tau) = \#(x,y) \;: \exists z_1,z_2,\dots,z_m :body(\tau) \land r(x, y)
\end{equation}
\textit{Standard Confidence (SC)} is the ratio of the number of head groundings to the body grounding in the KG.

\vspace{-4mm}
\begin{equation}
\label{eq:sc}
SC(\tau) = \frac{supp(\tau)}{\# (x^\prime, y^\prime) : \exists z_1,z_2,\dots,z_m :body(\tau)}
\end{equation}
\vspace{-4mm}
   
% \begin{itemize}
%     \item \textbf{Rule Support} : It 
%    
%     % \resizebox{.95\linewidth}{!}{$
%     % \displaystyle
%     % $}
%     % \item \textbf{Head Coverage} : The proportion of all ground pairs in KG with the head relation that are covered by the support of the rule.
%     % \[hc(\tau) = \frac{supp(\tau)}{\# (x^\prime, y^\prime) : r(x^\prime, y^\prime)}\]
    
%     \item 
% \end{itemize}

% In \cite{galarraga2013amie}, a partial rule is repeatedly extended by mining operators. Mining operators are essentially different forms of candidate atoms that can be added. At every step, partial rules are scored using the above metrics and output as a rule if it passes a certain threshold, extended by operators if the resulting partial rules has higher score that the previous else is rejected.  Repeated extension of partial rules is done by different forms of candidate atoms while are called mining operators.

\section{Motivation}
    \begin{figure}
    \centering
    \includegraphics[width=9.0cm]{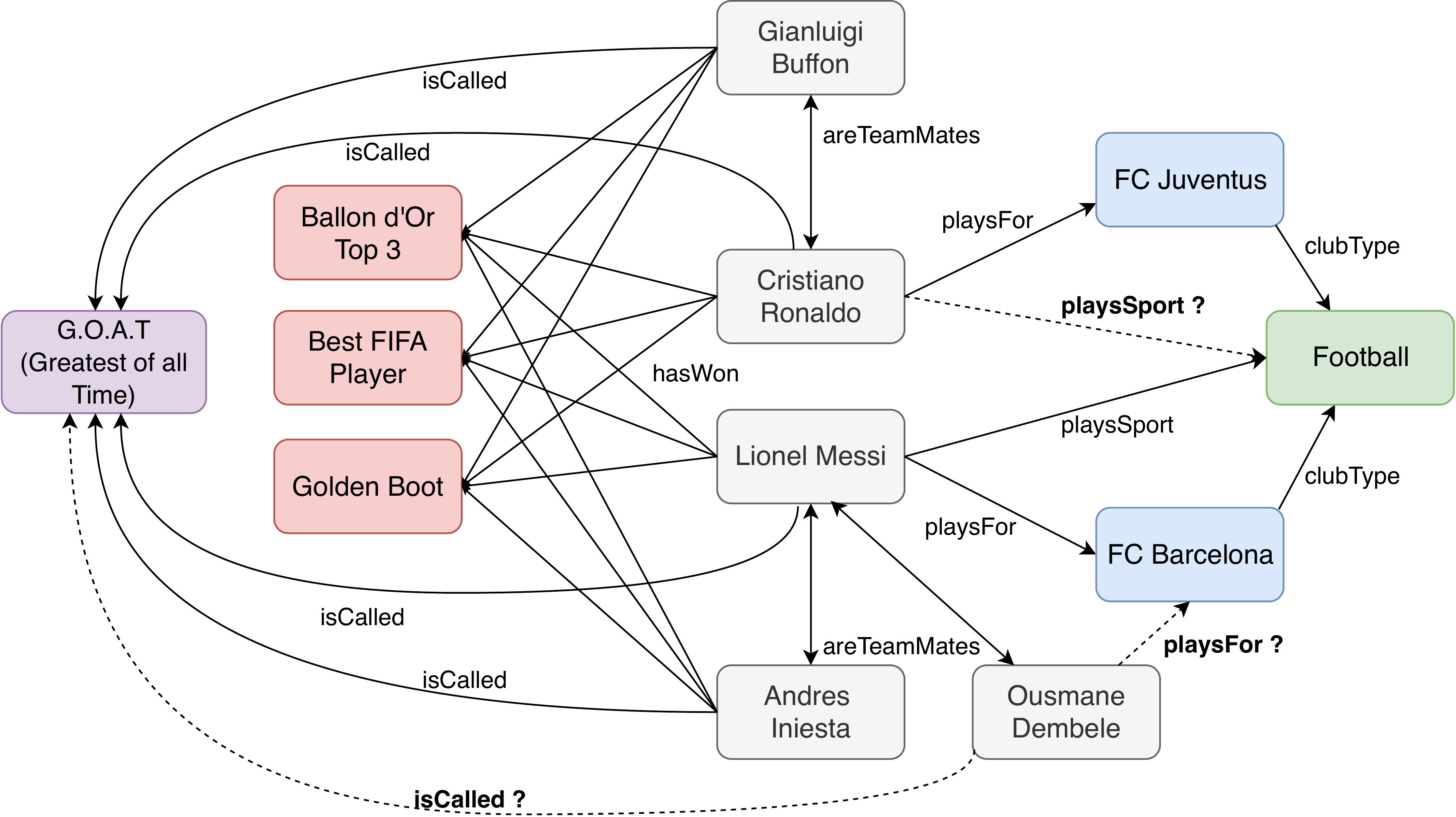}
    \caption{Entities and Relations in $G_f$. Known relationships shown as solid edges and possible relations in dashed.}
    
    \label{fig:sample_kg}
\end{figure}

Consider the sample $G_f$ in Fig. \ref{fig:sample_kg} drawn from a larger KG $G$ about sports. 
Suppose a rule learning system has found the following three rules: 

{
\small
\begin{align*}
&l_1: playsFor(V_1, V_3) \land  clubType(V_3, V_2) \rightarrow playsSport(V_1, V_2) \\
&l_2 : playsFor(V_1, V_2) \land playsFor(V_3, V_2) \rightarrow teamMate(V_1, V_3) \\
&l_3: isCalled(V_1, V3) \land teamMate(V_1, V_2) \rightarrow isCalled(V_2, V_3)
\end{align*}
}%

\noindent \textbf{Q1: Can the generalizing ability of logical rules help embedding methods learn better representations?} Rules $l_1$ and $l_2$ have a number of conforming instantiations in $G_f$ and one can argue for them to hold in the larger knowledge graph $G$. Say in a sample pertaining to basketball $G_b$, the entities are sparsely connected. Embedding representations of entities in such a ``less-connected" sub-graph will not be expressive because of poor data quality. On the other hand, rules $l_1$ and $l_2$ mined with support from facts in $G_f$ can accurately reason about entities in $G_b$. 

\noindent \textbf{Q2: Can feedback from an embedding model improve the quality of mined rules?} Embedding models are good at incorporating global patterns. For example, a possible explanation for Ronaldo being G.O.A.T (greatest of all time) is the presence of a large number of links between him and different sport awards (see Fig. \ref{fig:sample_kg}). An embedding method say, RESCAL \cite{rescal} can easily model the pattern: \textit{consistent top players of a sport are likely to be called G.O.A.T}. Specifically, the model can learn the features of ``consistent top player" and ``concept of eminence" for different entities from data.

{\small 
\[\mathbf{h}_{ronaldo} = \begin{bmatrix}
0.9\\
0.1
\end{bmatrix} \; \mathbf{M}_{isCalled} = \begin{bmatrix}
0.1 & 0.9\\
0.1 & 0.1
\end{bmatrix} \; \mathbf{t}_{G.O.A.T} = \begin{bmatrix}
0.2\\
0.9
\end{bmatrix} \]
}%

The first embedding feature of $\mathbf{h}$ and $\mathbf{t}$ might model ``consistent top player" and the second ``concept of eminence". Thus,
{\small
\[\phi \texttt{(ronaldo, isCalled, G.O.A.T)} = \mathbf{h}^{\top} \mathbf{M_r} \mathbf{t} = 0.76\]
}%
gives a high score denoting validity.

Now, although rule $l_3$ has multiple instantiations\footnote{(Buffon, Ronalado) and (Messi, Iniesta) are all G.O.A.T (see Fig. \ref{fig:sample_kg})} in $G_f$, it is quite noisy and unreliable. For e.g. it will incorrectly infer that Dembele (an upcoming young player) is G.O.A.T by the virtue of him being teammates with Messi. Now, such a noisy rule could have been pruned out if the information from an embedding method modelling \textit{consistent top player} were to be used. Specifically having,
{\small
\[\mathbf{h}_{dembele} = \begin{bmatrix}
0.3\\
0.1
\end{bmatrix}
\]
}%
\vspace{-1mm}
{\small
\[\phi \texttt{(dembele, isCalled, G.O.A.T)} = \mathbf{h}^{\top} \mathbf{M_r} \mathbf{t} = 0.26
\]
}%
gives a low score showing lower confidence compared to earlier. The two questions raised in the above examples motivates us to develop a cross feedback hybrid model for knowledge base reasoning.
\section{Related Work}
    % Survey articles \cite{nickel2015review} and \cite{wang2017knowledge} provide an excellent overview of embedding learning techniques. Methodology and review of rule mining methods for knowledge graphs can be found in \cite{galarraga2015fast}. 
Here we give a brief overview of embedding learning and rule mining methods and review relevant hybrid models from literature. 
\begin{figure*}
    \centering
    \includegraphics[width=15.0cm]{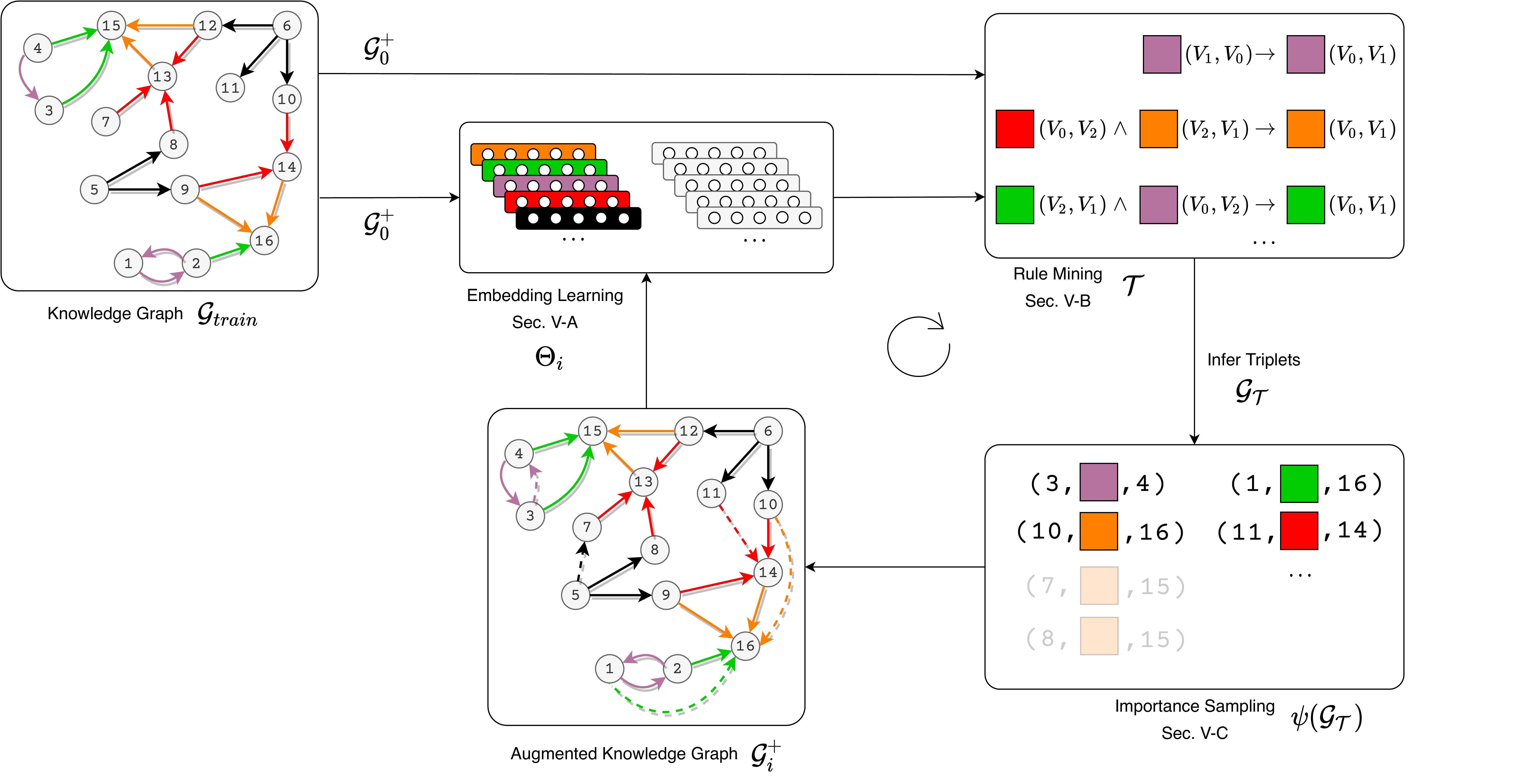}
    \caption{Overall framework architecture.}
    \label{fig:overall}
    \vspace{-5mm}
\end{figure*}
\subsection{Knowledge graph embeddings}
RESCAL \cite{rescal} one of the earliest works in KGE used a bilinear form for the scoring function inspired from matrix factorization. Specifically, $f(h,r,t) = \mathbf{h}^{\top} \mathbf{M_r} \mathbf{t}$  where entities $h$ and $t$ are represented by vectors in $\mathbb{R}^d$ and relations as matrices in $\mathbb{R}^{d\times d}$ that model pairwise interactions. DistMult \cite{distmult} simplifies RESCAL further by restricting $\mathbf{M_r}$ to be a diagonal matrix . This added simplicity does well for capturing symmetric relationships but bad for others. TransE \cite{transe} introduces entity and relation specific embeddings and models them as translations in vector spaces. ComplEx \cite{complex} is an extension to DistMult where embedding lie in $\mathbb{C}^d$. This accounts for order of the entities and thus is able to model asymmetric relations. A newer method RotatE \cite{rotate} is able to model symmetry, asymmetry, inversion and composition of relations. It uses the concept of rotations in complex space as opposed to translations in real space. Concretly, $f(h,r,t) = \left\Vert \mathbf{h} \circ \mathbf{r} - \mathbf{t}\right\Vert$ with constraint $|r_i| = 1$ where $\mathbf{h}, \mathbf{r}, \mathbf{t}  \in \mathbb{C}^d$ and $\circ$ is the element-wise product. ConvE \cite{conve} employs a 2D convolutional neural network to model the score function where the aim is to utilize the expressivity of multiple non-linear features in the architecture to better model relationships in the KG. A comprehensive survey of methods for Knowledge graph embeddings is given in \cite{nickel2015review} and  \cite{wang2017knowledge}.

\subsection{Rule mining}
These models assume that existence of a individual triples/facts can be inferred from observable features in the graph usually in the form of logical rules. The extracted rules are then used to infer new facts. Mining rules from a KB has its roots in inductive logic programming (ILP) \cite{muggleton1994inductive} and association rule mining \cite{agrawal1993mining} from the databases community. The use of declarative language biases \cite{ade1995declarative} help in restricting the large search space. Language biases like limiting rule length, having every atom to be transitively connected to each other in a rule (a.k.a \textit{connected} rules) and ensuring all variables in a rule to to be \textit{closed} (i.e. appears at least twice) offer a trade-off between the expressivity of rules and size of the search space. WARMR \cite{warmr} and its extension WARMER \cite{warmer} is based on the APRIORI algorithm \cite{agrawal1996fast} and makes use of a language bias where only conjunctive rules are mined. Sherlock \cite{sherlock} an unsupervised ILP learns first order Horn rules and infers new facts using probabilistic graphical models. To prune the search space it uses two heuristics:  statistical \textit{significance} and \textit{relevance}. All the above approaches are designed to work under the closed world assumption of KBs due to their treatment of the rule quality measure. AIME \cite{galarraga2013amie} and it's efficient version AIME+ \cite{galarraga2015fast} take into account the incompleteness of KBs by proposing a new quality measure based on partial completeness assumption. Additionally they mines Horn rules which are \textit{connected}, \textit{closed}, \textit{non-reflexive} and \textit{monotonic} (predicates in the rule body are all positive) rules. 

% \cite{gad2016exception} relaxes the \textit{monotonicity} bias and mines \textit{non-monotonic} rules under OWA. This enables their method to mine exception aware rules.

\subsection{Hybrid Methods}

The complementary strengths \cite{toutanova2015observed} of observed (rules, paths) and latent (embeddings) KG features  has given rise to a number of hybrid models in recent
years. 
% These methods broadly fall under three different paradigms: 1) Rules are used to aid embedding learning 2) Embedding methods are used in rule mining systems for pruning the search space; 3) Combined learning with iterative feedback between both methods. In the first two paradigms each methods is detached from the other and the third paradigm aims at learning both at the same time while making use of their complementary strengths.
In \cite{wang2015knowledge} link prediction is cast as an integer linear programming problem in which the objective function is from an embedding model and implication rules are used as constraints. This method is essentially a post processing step which helps in inference but not better embedding representations. Another paradigm of hybrid models is to perform some form of regularization on the embedding loss function using simple rules. In \cite{rocktaschel2015injecting} implication rules are first naively extracted by iterating over all possible relation pairs and then differentiable terms are added to the embedding objective function for each grounding of the extracted rules. The naive rule extraction process is costly and the procedure leads to a large number of regularization terms which doesn't scale. RUGE \cite{ruge} uses t-norm based fuzzy logic to model the rules and uses inferred triplets to perform embedding rectification. In \cite{ding2018improving} non-negativity constraints for entity embeddings and entailment constraint for relation embeddings is explored. \cite{fatemi2019knowledge} enforce the subsumption property by having an equality constraint $\mathbf{r_1} = \mathbf{r_2} - \mathbf{\delta}$, where $\mathbf{\delta}$ is a learnable non-negative vector that specifies how relation $r_1$ differs from $r_2$. 
% They also theoretically prove that this equality constraint can enforce subsumption in the learned embeddings. In both the above models, only one kind of simple rule constraint on the embeddings space is handled. \cite{gutierrez2018knowledge} propose a framework to study the effectiveness of KG embeddings in modelling ontological rules by viewing relations as regions in the embedding space. While they show how certain rules can be translated as constraints on convex regions they do not propose a practical method for learning.
Our work is closest to IterE \cite{zhang2019iterE} which is designed for improving embeddings of sparse entities. Seven ontology property axioms from OWL2 Web Ontology Language\footnote{https://www.w3.org/TR/owl2-primer/} are used to model relations in the KG. % These include $InverseOP(r_1, r_2)$,  $SymmetricOP(r)$, $TransitiveOP(r)$, $EquivalentOP(r_1, r_2)$ and $SubOP(r_1, r_2)$ (i.e. subsumption).
First a pool of valid axioms are generated, by randomly sampling $k$ triples and matching them to seven axiom templates. Then, inferred facts that relate to sparse entities help in learning better embeddings because they essentially provide extra information.
% The value of $k$ is proportional to the size of the search space so, a smaller $k$ will result in a small search space which might lead to low quality axioms. 
We consider horn rules that are more expressive and unlike them, we have no assumptions on the embedding score function. We improve embedding representations of both sparse and non-sparse entities. Different from constraint based hybrid methods, our selective augmentation of inferred facts provides 1) structure to the embedding space and 2) extra information for sparse entities.

\section{Our Approach}
    Here, we introduce our hybrid model for learning horn rules and vector space embeddings in a cross feedback paradigm. Initially, embedding learning is performed on the input KG resulting in entity and relation embeddings. The learnt embeddings are then used to guide the rule mining system. Further, the extracted rules are materialized and new inferred triplets are sampled for learning embeddings in the next iteration. Fig. \ref{fig:overall} shows the overall framework. In what follows, we describe the three main constituent parts (1) embedding learning, (2) rule learning with embedding feedback and (3) importance sampling of triplets.

% Different from standalone embedding learning methods which directly learn $\Theta$ from $G^+$, our method learns $\Theta$ from $G^+_{ext}$ an extension of $G^+$ . 

% where $G_{\mathcal{T}}$ is the set of inferred triplets derived from a set of learnt rules $\mathcal{T}$. The function $\psi$ is our sampling strategy defined in \ref{sec:sampling} to sample $G_{\mathcal{T}}$.

\subsection{Embedding Learning}
We associate a label $y_{hrt}$ to each triplet $(h,r,t)$ to model its truth value. Labels of triplets in $G^+_{i}$ are set to 1. As there are no negative triplet examples, we generate $G^-_{i}$ by negatively sampling $G^+_{i}$ and labels of triplets in $G^-_{i}$ are set to 0. A score function $\phi(h,r,t ; \Theta) : \mathcal{E} \times \mathcal{R} \times \mathcal{E} \rightarrow \mathbb{R}$ is used to measure the salience of a triplet $(h,r,t)$. We further map the output of the score function to a continuous truth value between $(0,1)$ using a sigmoid function $\sigma (z) = 1/(1+ exp(-z))$. 

\vspace{-2mm}
\begin{equation}
\label{eq:sigmoid_score}
    \xi(h,r,t) = \sigma (\phi(h,r,t;\Theta))
\end{equation}

The objective of embedding learning is to learn $\Theta_{i}$ for iteration $i$ by minimizing the loss over triplets in $G^+_{i}$ and $G^-_{i}$. $G_{i} = G^+_{i} \cup G^-_{i}$ represents our learning set for the current iteration $i$. Then,

\begin{equation}
\label{eq:min_loss}
    \min_{\Theta_{i}} \frac{1}{|G_{i}|} \sum_{((h,r,t),y_{hrt}) \in G_{i}} \mathcal{L} (\xi(h,r,t), y_{hrt})
\end{equation}

\noindent where $\mathcal{L}(x,y) = -y\log(x) - (1-y)\log(1-x)$ is the cross entropy loss between $x$ and $y$, $\xi(\cdot)$ is the function defined in Eq. \ref{eq:sigmoid_score}. It is important to note that in our method, learning is done on an extended set of rule enriched triples. Also, our method does not depend on specific or class of scoring functions unlike IterE \cite{zhang2019iterE}. All we require is that the score function $\phi$ output a real valued score. Table \ref{tab:score_functions} shows the score functions proposed by various state-of-the-art methods in the literature. 

\subsection{Rule Mining with Embedding Feedback}
In this step we aim to mine quality \textit{horn} rules from the set of initial triplets in $G^+_0$ using the current iteration embeddings $\Theta_i$.
% Our method utilizes the RuLES \cite{rules} framework which is based on the popular rule mining system AIME+
According to AIME+ \cite{galarraga2015fast}, rules are modelled as a sequence of atoms with the first atom as the head atom and other body atoms following it. The process of building a quality rule boils down to extending a partial rule sequence and carefully traversing the search space. Rule building is done by a set of mining operators that iteratively add atoms to partial rules until a termination criterion is met. To prune the search space during rule building filtering criteria is utilized. We now explain the three sub parts.

\subsubsection{Rule Building}
Initially, all possible binary atoms using relations in $\mathcal{R}$ are held in a priority queue. These represent partial rule heads with empty bodies. At each iteration of the algorithm, a single partial rule is dequeued and checked for \textit{termination} (defined below). If successful, it is output as a possible rule. If not, it is extended by each of the following operators:
\begin{itemize}
    \item Add a new dangling atom ($\mathcal{O}_D$) which uses a fresh variable for one argument and a shared variable/entity (used earlier in the rule) for the other. 
    \item Add a new instantiated atom ($\mathcal{O}_I$) which uses a shared variable/entity for one argument and an entity for the other.
    \item Add a new closing atom ($\mathcal{O}_C$) which uses shared variable/entity for both arguments.
\end{itemize}
The expansion produces multiple candidate rules. All candidate rules are checked if they can be pruned by \textit{filtering criteria}. Pruned ones are discarded and non pruned ones are enqueued for the next iteration. The iterative algorithm is run until the queue is empty.

\subsubsection{Filtering Criteria}
\label{sec:filtering}
The application of mining operators to a rule produces a set of candidate rules. Since not all candidate rules are promising, this steps aims to filter some of them. Ideally, filtering criteria should allow the generation of rules that (1) explain facts in the KG and (2) infer facts outside the observable KG while being consistent w.r.t global ``soft" KG properties. Classical measures of rule quality like \textit{standard confidence} and \textit{PCA confidence} introduced by \cite{galarraga2013amie} are based on the observable KG. They work well in selecting rules that explain known facts. Every so often, candidate rules infer facts that are outside the known KG and classical measures neglect them. We seek to use the learned embedding as a proxy for measuring the quality of candidate rules which infer facts outside the known KG. This is based on the intuition that embeddings $\Theta$ can capture global statistical patterns. To this end, given a candidate rule, we average the individual embedding scores of all its newly inferred atoms. We call this measure \textit{embedding confidence (EC)}. Given a candidate rule $\tau$, consider the set $S_\tau$ defined in Eq. \ref{eq:s_t}. The unobserved facts predicted by instantiating $\tau$ is given by $S_{\tau} \setminus G^+_{0}$. Then, 
\begin{equation}
\small
    \begin{split}
        EC_{}(\tau) &= \frac{1}{|S_{\tau} \setminus G^+_{0}|}\sum_{(h,r,t) \in S_{\tau} \setminus G^+_{0}} \xi(h,r,t)
    \end{split} \label{eq:emb_measure}
\end{equation}
where $\xi(\cdot)$ is the function defined in Eq. (\ref{eq:sigmoid_score}). $EC \in (0,1)$ as $\xi(\cdot)$ is designed to give confidence values between $(0,1)$. We consider the weighted average of the classical standard confidence i.e. $SC$ (Eq. \ref{eq:sc}) and our embedding confidence $EC$ (Eq. \ref{eq:emb_measure}) for the final rule quality measure $Q$. 
\begin{equation}
\small
\label{eq:rule_quality_measure}
    Q(\tau) = (1 - \omega) SC(\tau) + \omega EC(\tau)
\end{equation}
where $\omega$ the weight factor is a model hyper-parameter. Thus, for each candidate rule we check if there is an increase w.r.t rule quality measure $Q$ and if not it is discarded.
Before calculating the $Q$ score for a candidate rule, we use a heuristic pruning strategy followed by AIME+ \cite{galarraga2013amie} to discard non-interesting rules. This is done so as to reduce the load on calculating the $Q$ score. One quick statistical check we perform is to make sure a candidate rule covers more than 1\% of known facts using \textit{head coverage} \cite{galarraga2013amie}. We also incorporate \textit{language biases} like restricting the search to rules of length three in order to deal with the vast search space.

\noindent\textbf{Termination Criteria }
As the extension operators produce rules that are not necessarily \textit{closed}, we follow \cite{galarraga2013amie} in outputting only closed and connected rules. 

\subsection{Importance Sampling of Triplets}
\label{sec:sampling}
Once the rule mining system has extracted rules with a our quality measure, we now incorporate them in refining the embeddings for the next iteration. To do so, the first task is to select top-K rules by ranking the extracted rules according to their quality measure $Q$ which is already calculated (Eq. \ref{eq:rule_quality_measure}). %\jn{Still need to state what value of K you used, or add it as a parameter in your algorithm and state in experimental details.} 
Let $\mathcal{T}$ represent the set of all the selected rules. Then, the newly inferred triplets from the rules in $\mathcal{T}$ will be represented by,

\vspace{-4mm}
\begin{equation}
\small
\label{eq:G_t}
    G_\mathcal{T} = ( \bigcup\limits_{\tau \in \mathcal{T}} S_{\tau}) \setminus G^+_{0}
\end{equation}
where $S_{\tau}$ (Eq. \ref{eq:s_t}) is the set of inferred triplets from rule $\tau$. Next we augment the current positive triplet set $G^+_{i}$ with $G_\mathcal{T}$ to create $G^+_{i+1}$ which will be used for embedding learning in iteration $i+1$. Instead of naively augmenting all of $G_\mathcal{T}$ with label 1, we use an \textit{importance sampling} scheme that uses $\Theta_i$ from current iteration $i$. Specifically, we sample triplets from the following distribution,
\begin{equation}
\small
    p((h,r,t)| G_{\mathcal{T}}) = \frac{\exp{\beta \phi(h,r,t; \Theta_i)}}{\sum\limits_{((h_j, r_j, t_j) \in G_{\mathcal{T}})} \exp \beta \phi(h_j, r_j, t_j; \Theta_i) }
\end{equation}
where $\phi(\cdot)$ is the embedding score function and $\beta$ is the sampling temperature. %We treat the above probability as the weight for each triplet that will be augmented so, its label is set using the above equation. Note that the labels for triplets in $G_\mathcal{T}$ will have a value in $(0,1)$. Thus $y_{hrt}  = p((h,r,t) | G_{\mathcal{T}})$. %JN: I think this part of the description is not needed. It just adds confusion.
The set of input triplets for the next embedding learning iteration is generated as shown in line \ref{algo:augment} of  Algorithm \ref{algo}, where $\psi$ represents the importance sampling strategy.

\begin{algorithm}
\setstretch{1.2}
\caption{Hybrid Learning Procedure}
\label{algo}
\textbf{Input}: Given initial training knowledge graph ($\mathcal{KG}$), embedding method $\phi$   \\
\textbf{Output}: Final embeddings $\Theta$ and mined rules $\mathcal{T}$
\begin{algorithmic}[1]
\State $G^+_{0} \gets \{((h,r,t), y_{hrt}=1) | (h,r,t) \in \mathcal{KG}\}$
\State{$\mathcal{T} \gets \emptyset$}
\State{Randomly initialize entity and relation embeddings $\Theta_{0}$}
\For{$i = 1 : M$}
\State{$G^-_{i-1} \gets NegativeSampling(G^+_{i-1})$}
\State{$\Theta_{i} \gets EmbeddingLearning(G^+_{i-1}, G^-_{i-1})$} \Comment{Eq. \ref{eq:min_loss}}
\State{$\mathcal{T} \gets RuleMining(G_{0}^+ , \Theta_{i})$}
\State{$ G^+_{i} = G^{+}_{i-1} \; \cup \; \psi(G_{\mathcal{T}})$} \Comment{Sec. \ref{sec:sampling}} \label{algo:augment}
\EndFor
\State \textbf{return} $\Theta_{M}, \mathcal{T}$ \end{algorithmic}
\end{algorithm}

\section{Experiments and Analysis}
% \usepackage{booktabs}
% \usepackage{multirow}
\begin{table*}[ht]
\centering
\caption{Benchmark results for link prediction. Comparison against state-of-the-art standalone KGE and hybrid methods. Results of [$\spadesuit$] are taken from \protect\cite{rotate}. YAGO3-10 results for methods with [$\bigstar$] are taken from \protect\cite{ruge}. Results of [$\clubsuit$] are produced by us using code provided by respective authors. All other results are taken from corresponding original papers. Best scores are in \textbf{bold}, runner-up is \underline{underlined} and ``$*$" represents statistically significant improvements over RotatE (paired t-test; p-value $< 0.01$)}
\small
\renewcommand{\arraystretch}{1.1} % Default value: 1
\begin{tabular}{lcccc|cccc|cccc}
\toprule
\multirow{2}{*}{} & \multicolumn{4}{c|}{FB15K-237} & \multicolumn{4}{c|}{WN18RR}  & \multicolumn{4}{c}{YAGO3-10}\\ \cmidrule(l){2-13} & MRR & Hit@1 & Hit@3 & Hit@10 & MRR & Hit@1 & Hit@3 & Hit@10 & MRR & Hit@1 & Hit@3 & Hit@10\\ \midrule
TransE [$\spadesuit, \bigstar$ ] & 0.293 & 0.140 & 0.268 & 0.463            & 0.226 & - & - & 0.501                 &0.303  & 0.218 & 0.336 &  0.475 \\ 
DistMult [$\spadesuit$] & 0.241 & 0.155 & 0.263 & 0.419          & 0.430 & 0.390 & 0.440 & 0.490       & 0.340 & 0.240  & 0.380  & 0.540 \\
ComplEx [$\spadesuit$] & 0.247 & 0.158 & 0.275 & 0.428           & 0.440 & 0.410 & 0.460 &  0.510      & 0.360 & 0.260 &  0.400 & 0.550\\ 
ConvE  & 0.325 & 0.237 & 0.356 & 0.501             & 0.430 &  0.400& 0.440 &  0.520      & 0.440  & 0.350 & 0.490 & 0.620 \\ 
RotatE  & \underline{0.338} & \underline{0.241} & \underline{0.375} & \underline{0.533}            & \underline{0.476} & \textbf{0.428} & \textbf{0.492} &  \underline{0.571}      &  \underline{0.495} & \underline{0.402} & \underline{0.550} & \underline{0.670}\\ \midrule
% KALE [$\star$] &  - & - & - & -      &  -  & -  & - & -                               & 0.321 & 0.215 & 0.372 & 0.522\\
RUGE [$\clubsuit, \bigstar$] & 0.169 & 0.087 & 0.181 & 0.345            & 0.231 & 0.218 & 0.387 &  0.439      & 0.431 & 0.340 & 0.482  & 0.603\\
NNE-AER [$\clubsuit$] & 0.317 & 0.183 & 0.294 & 0.478           & 0.431 & 0.412 & 0.437 & 0.467                 & 0.390 & 0.310 & 0.419 & 0.597\\
% IterE & \\
pLogicNet &  0.332 & 0.237 & 0.369 & 0.528      &  0.441  & 0.398  & 0.446 & 0.537 & - & - & - & -\\ \midrule
Our method & $\textbf{0.478}^*$ & $\textbf{0.409}^*$  & $\textbf{0.522}^*$ & $\textbf{0.624}^*$ & \textbf{0.479} & \underline{0.425} & \underline{0.489}  & \textbf{0.579}
& $\textbf{0.557}^*$ & $\textbf{0.469}^*$& $\textbf{0.621}^*$ & $\textbf{0.748}^*$ \\ \bottomrule
\end{tabular}
\label{tab:benchmark_results}
\end{table*}

\begin{table}
\centering
\caption{Statistics of Datasets}
\small
\small
\begin{tabular}{lccccc}
\toprule
Dataset & $|\mathcal{E}|$ & $|\mathcal{R}|$ & \#Train & \#Valid & \#Test \\ \midrule
% FB15K & 14,951 & 1,345 & 483,142 & 50,000 & 59,071 \\
% WN18 & 40,943 & 18 & 141,442 & 5000 & 5000 \\
FB15K-237 & 14,541 & 237 & 272,115 & 17,535 & 20,466 \\
WN18RR & 40,943 & 11 & 86,835 & 3,034 & 3,134 \\ 
YAGO3-10 & 123,182 & 37 & 1,079,040 & 5,000 & 5,000 \\ \midrule
\begin{tabular}[c]{@{}l@{}}FB15K-237-\\ sparse\end{tabular} & 14,541 & 237 & 272,115 & 10,671 & 12,454 \\
\begin{tabular}[c]{@{}l@{}}WN18RR-\\ sparse\end{tabular} & 40,943 & 11 & 86,835 & 1,609 & 1,661 \\ \bottomrule
\end{tabular}

\label{tab:datasets}
\end{table}
\noindent \textbf{Data} We evaluate our proposed method on multiple widely used benchmark datasets. FB15k-237 \cite{toutanova2015observed}, which has facts about movies, actors, awards, sports and sport teams, is a subsset of FB15k \cite{transe} with no inverse relations. WN18RR \cite{conve} a subset of WordNet\footnote{WordNet - https://wordnet.princeton.edu/} is a KG describing lexical relations between words. YAGO3-10 \cite{mahdisoltani2015yago3} deals with descriptive attributes of people. We also consider sparse variants from \cite{zhang2019iterE} for one of our experiments. Statistics of all datasets are given in Table \ref{tab:datasets}. We use the original train/valid/test splits provided by authors of the dataset and represent it as $G_{train}, G_{valid}$ and $G_{test}$.
% \usepackage{booktabs}
% \usepackage{multirow}
\begin{table*}[ht]
\centering
\caption{Sparse KG results. Results of [$\clubsuit$] are produced by us using code provided by respective authors. Other results are taken from \protect\cite{zhang2019iterE}. Best scores are highlighted in \textbf{bold} and runner-up is \underline{underlined}.}
\small
\renewcommand{\arraystretch}{1.0} % Default value: 1
\begin{tabular}{lcccc|cccc}
\toprule
\multirow{2}{*}{} & \multicolumn{4}{c|}{FB15K-237-sparse} & \multicolumn{4}{c}{WN18RR-sparse} \\ \cmidrule(l){2-9} & MRR & Hit@1 & Hit@3 & Hit@10 & MRR & Hit@1 & Hit@3 & Hit@10\\ \midrule
TransE   & 0.238 & 0.164 & 0.261 & 0.385           & 0.146 & 0.034 & 0.247 & 0.288 \\
DistMult  & 0.204 & 0.128 & 0.226 & 0.362            & 0.255 & 0.238 & 0.260 & 0.225\\
ComplEx  & 0.197 & 0.120 & 0.217  & 0.354         & 0.259 & 0.246 & 0.262 & 0.286 \\
RotatE [$\clubsuit$] & \underline{0.292} & \underline{0.213} & \underline{0.320}  & \underline{0.445}         & \underline{0.340} & \underline{0.299} & \underline{0.354} & \underline{0.422} \\ \midrule

RUGE [$\clubsuit$]& 0.241 & 0.172 & 0.267  & 0.393         & 0.145 & 0.199 & 0.263 & 0.292 \\
NNE-AER [$\clubsuit$] & 0.212 & 0.133 & 0.238  & 0.395         & 0.279 & 0.284 & 0.304 & 0.305 \\

IterE + axioms & 0.247 & 0.179 & 0.262 & 0.392            & 0.274 & 0.254 & 0.281 & 0.314 \\ \midrule
Our method & \textbf{0.401} & \ \textbf{0.322} & \textbf{0.435}  & \textbf{0.547} & \textbf{0.350} & \textbf{0.312} & \textbf{0.364} & \textbf{0.448} \\ \bottomrule
\end{tabular}
\label{tab:sparse_results}
\end{table*}

\noindent\textbf{Evaluation Protocol}
We consider the standard link prediction task following a standard protocol introduced by \cite{transe}. For a test triplet $(h, r, t)$, we replace the head $h$ with each entity $e_i \in \mathcal{E}$ and find the score of $(e_i, r, t)$ using $\phi$. From the descending order of scores, the rank of the correct entity i.e. $h$ is found. This gives us the head rank. Similarly, we run the same procedure for the tail $t$ that gives the tail rank. Finally, the average of head and tail ranks is used to 
% Finally, we average the two to obtain $\mathfrak{R} = \frac{1}{2} (\mathfrak{R}_h + \mathfrak{R}_t)$. A lower $\mathfrak{R}$ means that the model scored the test triple higher which is desired. The average of all individual test $\mathfrak{R}$'s gives the \textit{Mean Rank (MR)} for the whole test set. 
report popular metrics like \textit{Mean Reciprocal Rank (MRR)} and percentage of predicting ranks within N which is \textit{Hit@N}. Higher values in both metrics signify better results. Also, during the ranking process, we make sure that the replaced triplet does not exist in either the training, validation, or test set. This corresponds to the ``filtered” setting in \cite{transe}. \cite{sun2019re} propose a RANDOM protocol to be used when choosing triplets that get the same score by the model. Following such a protocol, when we generate the rankings and if more than two entities receive the same score, we randomly pick one of them because picking in any order will be unfair as argued by \cite{sun2019re}.

\noindent \textbf{Hyperparameter Setting}
We fine-tune the hyper-parameters on the validation set $G_{valid}$. We grid searched embedding dimension $d$ in $\{200, 500, 1000\}$, batch size $\mathcal{B}$ in $\{256, 512, 1024\}$, embedding weight factor $\omega$ in $\{0.1, 0.3, 0.5, 0.7, 0.9\}$ and top-K extracted rules for $K$ in $\{50, 100, 200, 500\}$. Initially, all embeddings are randomly initialized and the cross feedback learning procedure is run with $M$ set to 10. (see line 4 in Algo. \ref{algo}). During each global iteration, embedding learning is run for $k$ steps. We set $k$ to 100. During rule mining, we mine rules with length of at most 3 to keep a check on the search space. $G_{valid}$ is used for early stopping and obtaining the best model on which $G_{test}$ is evaluated.

\subsection{Embedding Evaluation}

We compare our model with state-of-the-art standalone knowledge graph embedding methods representing a variety of modelling approaches, viz. TransE \cite{transe}, DistMult \cite{distmult}, ComplEx \cite{complex}, ConvE \cite{conve} and RotatE \cite{rotate}. The standalone models rely only on observed triplets in KG and use no logical rules. We further compare with additional hybrid baselines, including RUGE \cite{ruge} and NNE-AER \cite{ding2018improving} which make use of certain logical rules to constrain the vector space of $\Theta$ and pLogicNet \cite{qu2019probabilistic} which utilizes probabilistic logical rules in a Markov Logic Network (MLN) framework. For all baseline models we run (marked with $\clubsuit$ in result table), optimal hyper-parameters are obtained using grid search.  
Table \ref{tab:benchmark_results} shows the evaluation on different datasets. We can see that our model outperforms both standalone and hybrid baselines by a large margin for FB15K-237 and YAGO3-10 datasets. Both the two datasets have diverse relation patterns like composition, symetry and inversion in them and mined rules can infer rich triplets which leads to increased data quality when augmented to the original KG. There is no statistically significant improvement for WN18RR because the number of rules that are mined from WN18RR are only a few tens whereas for FB15K-237 and YAGO3-10 they are in the hundreds. Number of rules mined is directly correlated to the number of relations in the dataset (see Table \ref{tab:datasets}). Moreover, WN18RR dataset is less diverse and mainly made up of relations that conform to symmetry pattern. Constraint based hybrid approaches perform poorly across dataset as they don't model all rule patterns. We also perform a paired t-test between our method and the best baseline i.e. RotatE with a p-value $<$ 0.01 using 5 different trials to show statistical significance.

\subsection{Rule Evaluation}
\textit{Precision} measures the ability of the rules to infer true facts beyond the train set. Mathematically, given test set $G_{test}$, the precision of a set of mined (using $G_{train}$) rules $\mathcal{T}$ is,

\vspace{-2mm}
\begin{equation}
\small
\label{eq:precision}
    precision(\mathcal{T}) = \frac{|G_{\mathcal{T}} \cap G_{test}| }{|G_{\mathcal{T}}|}
\end{equation}
$G_{\mathcal{T}}$, defined in Eq. (\ref{eq:G_t}) is the set of all newly inferred triplets from the set of rules $\mathcal{T}$. Using this metric, we compare against AIME+ \cite{galarraga2015fast}. We compare the top-k rules that are output for better understanding. Results (Table \ref{tab:rule_eval__results}) show that our embedding confidence measure when combined with standard filtering techniques used by AIME+ leads to higher quality rules. We conclude that feedback from embeddings that carry ``global" KG pattern information is useful when mining rules. 

% \usepackage{booktabs}
% \usepackage{multirow}
\begin{table}[ht]
\centering
\small
\caption{Quantitative evaluation of learned top-k rules using $precision$ (Eq.\ref{eq:precision}). AIME+ uses \textit{PCA confidence} while our method utilizes \textit{Embedding confidence} (Eq. \ref{eq:rule_quality_measure}) for measuring rule quality.}
\renewcommand{\arraystretch}{1.1} % Default value: 1
\begin{tabular}{ccc|cc}
\toprule
\multirow{2}{*}{top-K} & \multicolumn{2}{c|}{FB15K-237} & \multicolumn{2}{c}{WN18RR} \\ \cmidrule(l){2-5} & AIME+ & Our Method & AIME+ & Our Method \\ \midrule
10  &  0.357 & \textbf{0.614}        & 0.369 & \textbf{0.375}        \\
20  & 0.422 & \textbf{0.693}           & 0.162 & \textbf{0.194}        \\ 
50  & 0.585 & \textbf{0.712}            & - & -         \\ 
100 & 0.613 & \textbf{0.761}            & - & -         \\
200 & 0.679 & \textbf{0.785}            & - & -         \\
500 & 0.384 & \textbf{0.662}             & - & - \\
\bottomrule
\end{tabular}
\label{tab:rule_eval__results}
\end{table}

\subsection{Training with Different Score Functions}
One may argue that our boost in performance is from a particular score function we used for embedding learning (i.e. RotatE). We experimentally show in Table \ref{tab:kge_score_function_results} that our cross feedback learning approach significantly improves performance when used with different scoring functions. The point we make here is that our rich data augmentation approach brings in deterministic structure leading to better embedding representations irrespective of the scoring function used for embeddings.

% \usepackage{booktabs}
% \usepackage{multirow}
\begin{table}[h]
\centering
\caption{Performance of our cross feedback paradigm with different embedding learning methods (i.e. $\phi$ is from Table \ref{tab:score_functions}). Results in the second row for each embedding method, i.e. with brackets, report performance without our paradigm.}
\small
\renewcommand{\arraystretch}{.9} % Default value: 1
\begin{tabular}{lcc|cc}
\toprule
\multirow{2}{*}{\begin{tabular}[c]{@{}l@{}}Embedding\\ Methods \end{tabular}} & \multicolumn{2}{c|}{FB15K-237} & \multicolumn{2}{c}{YAGO3-10} \\  \cmidrule(l){2-5} & MRR & Hit@10 & MRR & Hit@10 \\ \midrule
TransE  & \begin{tabular}[c]{@{}c@{}}\textbf{0.466}\\ (0.293) \end{tabular}  & \begin{tabular}[c]{@{}c@{}}\textbf{0.595}\\ (0.42)\end{tabular}         

& \begin{tabular}[c]{@{}c@{}}\textbf{0.436}\\ (0.303)\end{tabular}  & \begin{tabular}[c]{@{}c@{}}\textbf{0.619}\\ (0.475)\end{tabular}        

\\ \midrule
ComplEx  & \begin{tabular}[c]{@{}c@{}}\textbf{0.401}\\ (0.247)\end{tabular}  & \begin{tabular}[c]{@{}c@{}}\textbf{0.570}\\ (0.428)\end{tabular}               

& \begin{tabular}[c]{@{}c@{}}\textbf{0.501}\\ (0.417)\end{tabular}  & \begin{tabular}[c]{@{}c@{}}\textbf{0.679}\\ (0.603)\end{tabular}  
\\  \midrule
RotatE & \begin{tabular}[c]{@{}c@{}}\textbf{0.478}\\ (0.338)\end{tabular}  & \begin{tabular}[c]{@{}c@{}}\textbf{0.624}\\ (0.533) \end{tabular}             

& \begin{tabular}[c]{@{}c@{}}\textbf{0.557}\\ (0.495)\end{tabular}  & \begin{tabular}[c]{@{}c@{}}\textbf{0.748}\\ (0.670)\end{tabular}           
\\
\bottomrule
\end{tabular}

\label{tab:kge_score_function_results}
\end{table}

\subsection{Sparse KG Evaluation}
Here we compare our model against IterE \cite{zhang2019iterE} for the sparse KG setting using the same datasets provided by the authors of IterE. In the sparse versions of FB15K-237 and WN18RR, only entities with \textit{sparsity} (Eq. \ref{eq:sparsity} ) greater than 0.995 are allowed in the validation ($G_{valid}$) and test ($G_{test}$) sets while training sets ($G_{train}$) remain unchanged. 
According to \cite{zhang2019iterE}, sparsity of an entity $e$ is defined as,
\begin{equation}
    \label{eq:sparsity}
    sparsity(e) = 1 - \frac{freq(e) - freq_{min}}{freq_{max} - freq_{min}}
\end{equation}
where freq(e) is the frequency of entity $e$ participating in a triple among all triples in $G_{train}$. $freq_{min}$ and $freq_{max}$ are minimum and maximum frequency of all entities in $\mathcal{E}$.

These datasets provides us a way to evaluate if our model has the ability to improve embedding representations of sparse entities. From the results (Table \ref{tab:sparse_results}), it is clear that our method significantly improves the score on all metrics for both sparse versions compared to baselines. Embeddings guiding the search space of the rule mining system instead of a naive fixed $k$ pruning strategy used by IterE and the importance sampling mechanism employed by us leads to effective utilization of the generalizing power of rules. IterE is tied to a certain class of embedding methods which assume a linear map between subject and object entities while our model places no such assumptions. RotatE, the state-of-the-art standalone model also suffers when reasoning about sparse entities. Although RUGE and NNE-AER do better than some standalone methods, they fall short when compared to a our method.

\section{Additional Analysis}
    \subsection{Computational Complexity}
With regard to embedding learning, our approach has the same complexity as the score function that is chosen. If RotatE is used, the space complexity is $\mathcal{O} (d(|\mathcal{E}| + |\mathcal{R}|))$ which scales linearly w.r.t number of entities, number of relations and  embedding dimension. Each iteration of our learning procedure has a time complexity of $\mathcal{O}(b + kbd + \bar{n}d)$ for negative sampling, embedding learning and importance sampling combined where b represents batch size, d is embedding dimension, $k$ the inner embedding learning epochs (used by Eq. \ref{eq:min_loss}) and finally $\bar{n}$ is average size of $G_{\mathcal{T}}$ (see Eq. \ref{eq:G_t}) for each iteration. As $k$ is usually set to a small number and usually the number of average entailments is fewer than batchsize i.e. $\bar{n} < b $ our time complexity is on par with conventional KGE methods which have a complexity of $\mathcal{O} (bd)$. It is important to note that the above embedding time complexity does not depend on the size of the input graph $G_{i}$ at iteration $i$ because training is done using SGD in minibatch mode. As we also run rule mining with embedding guidance every global iteration it adds additional complexity. AIME+ \cite{galarraga2015fast} is known to be very efficient with its use of an in-memory database to store the KG and its use of various optimizations to prune the large search space.

\subsection{Iterative Learning Profile}
\begin{figure}[h]
    \centering
    \begin{subfigure}[t]{0.5\textwidth}
        \centering
        \includegraphics[width=0.75\linewidth]{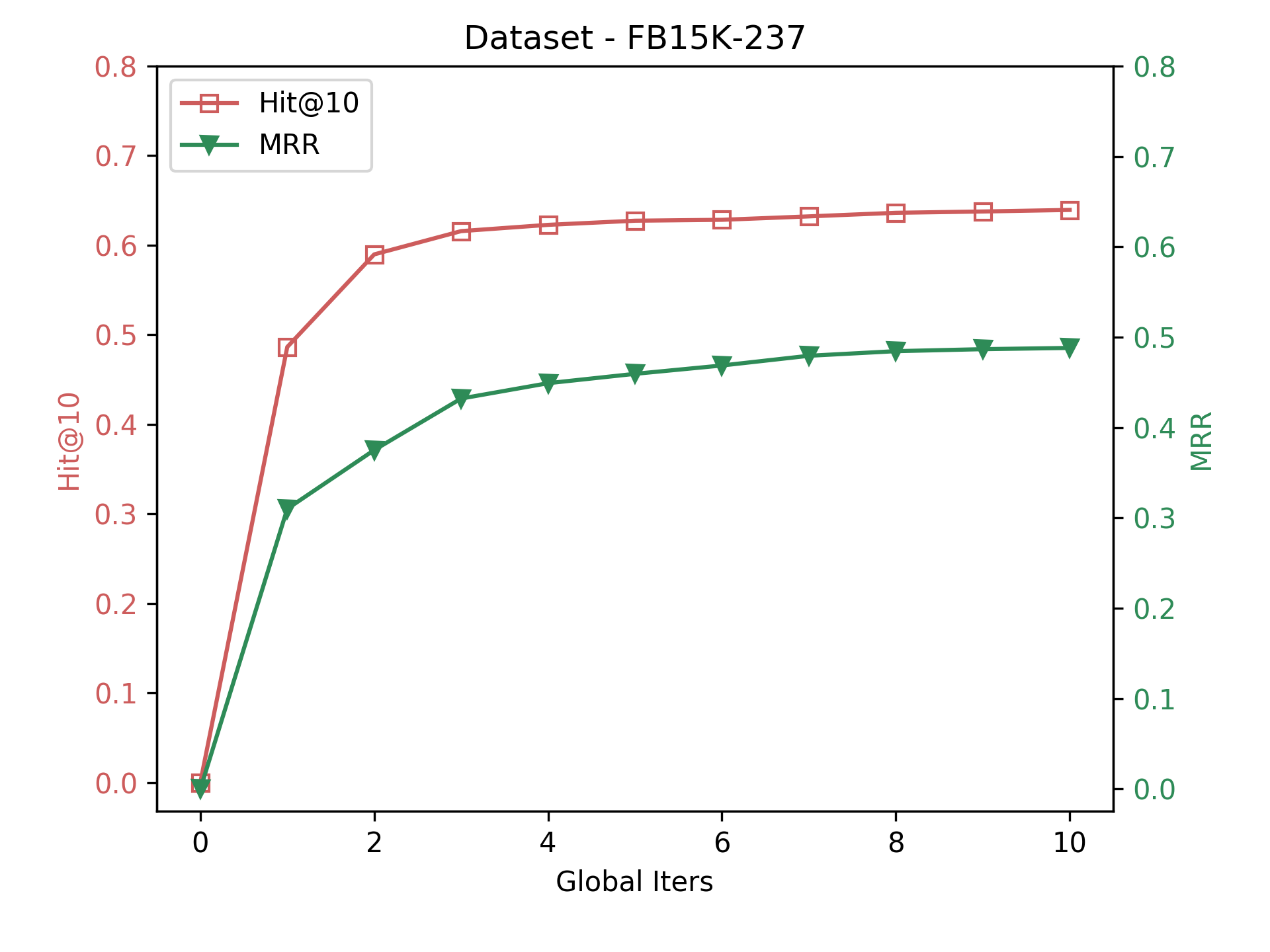}
    \end{subfigure}%
    ~ 
    
    \begin{subfigure}[t]{0.5\textwidth}
        \centering
        \includegraphics[width=0.75\linewidth]{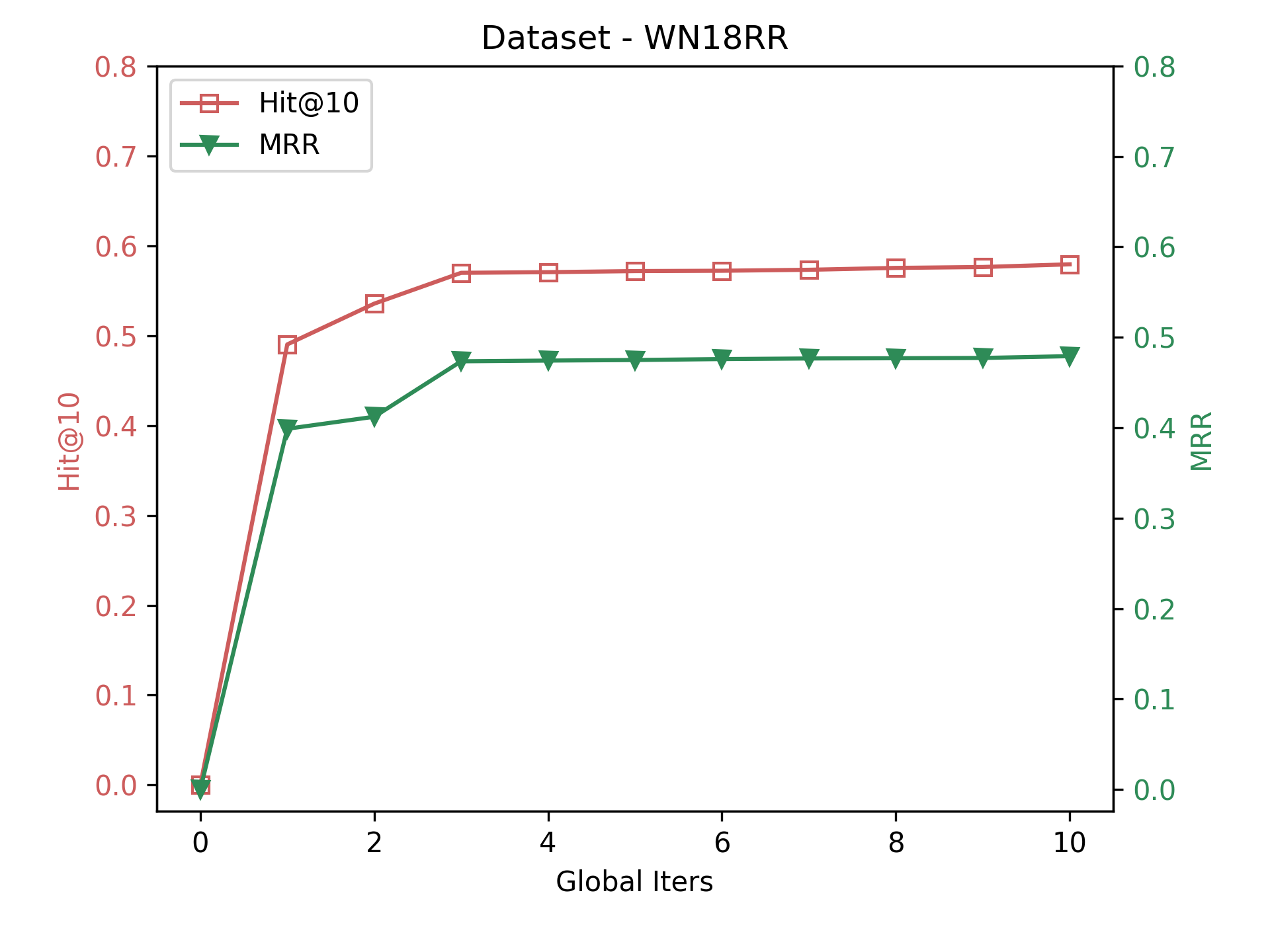}
    \end{subfigure}
    \caption{Link Prediction Performance vs Global Iterations}
    \vspace{-4mm}
    \label{fig:perf_vs_iters}
\end{figure}

Here we assess how performance (e.g., Hit@10, MRR) varies with the number of gloabl iterations. Fig. \ref{fig:perf_vs_iters} shows our findings for FB15K-237 and WN18RR. Big gains are observed after two iterations and performance continues to increase until it levels off around iteration eight. These plots indicate that these is a positive effect when we sample triplets from rule mining and augment them for embedding learning. 

\subsection{Case Study}

% \usepackage{booktabs}
% \usepackage{graphicx}
\begin{table*}[ht]
\centering
\small
\caption{Qualitative results showing link prediction rank change of entities of test triplets by comparing our method and baseline RotatE. The associated rule used for providing feedback is shown for each test triplet}
\renewcommand{\arraystretch}{1.2}
\resizebox{\textwidth}{!}{%
\begin{tabular}{@{}ll@{}}
\toprule
Test Triplet & (\texttt{Louis Armstrong, /people/person/nationality, United States})                                           \\
Rule         &  /music/artist/origin($V_0, V_2$) $\land$ /location/administrative\_division/country($V_2, V_1$) $\rightarrow$ /people/person/nationality($V_0, V_1$)\\
Rank Change  & \begin{tabular}[c]{@{}l@{}}head: 1154 $\rightarrow$ 68 (+1086) \\ tail: 1 $\rightarrow$ 1 (+0)\end{tabular}                  \\ \midrule

Test Triplet & (\texttt{15 Minutes, /film/country, United States})                                           \\
Rule         & /film/production\_companies($V_0, V_2$) $\land$ /organization/headquarters./location/country($V_2, V_1$) $\rightarrow$ /film/country($V_0, V_1$)\\
Rank Change  & \begin{tabular}[c]{@{}l@{}}head: 314 $\rightarrow$ 12 (+302) \\ tail: 1 $\rightarrow$ 1 (+0)\end{tabular}                  \\ \midrule

Test Triplet & (\texttt{Poland, /location/second\_level\_divisions, Szczecin })                                           \\
Rule         & /bibs\_location/country($V_2, V_0$) $\land$ /location/county\_place/($V_2, V_1$)$\rightarrow$ /location/second\_level\_divisions($V_0, V_1$)\\
Rank Change  & \begin{tabular}[c]{@{}l@{}}head: 1 $\rightarrow$ 1 (+0) \\ tail: 384 $\rightarrow$ 8 (+376)\end{tabular}                  \\ \midrule

Test Triplet & (\texttt{Fort Lauderdale, /location/location/time\_zones, Eastern Time Zone})                                           \\
Rule         & /location/location/contains($V_0, V_2$) $\land$ /location/location/time\_zones($V_2, V_1$) $\rightarrow$ /location/location/time\_zones($V_0, V_1$)\\
Rank Change  & \begin{tabular}[c]{@{}l@{}}head: 414 $\rightarrow$ 45 (+369)\\ tail: 1 $\rightarrow$ 1 (+0)\end{tabular}                  \\ \bottomrule
\end{tabular}%
}

\label{tab:eg_triplets}
\end{table*}
We give some qualitative examples from FB15K-237 dataset to demonstrate the effectiveness of our model. First, consider Table \ref{tab:eg_triplets}. It shows multiple test triplets from $G_{test}$ and their corresponding head and tail rank change when comparing RotatE \cite{rotate} as baseline with our method. The relevant rule that provides feedback to embedding learning in our framework is also shown for each test triplet. As an example, take the first test triplet (\texttt{Louis Armstrong, /people/person/nationality, United States}). Because $\texttt{Louis Armstrong}$ is a sparse entity with sparsity 0.995, traditional embedding methods suffer in the head prediction task i.e. asking \texttt{(?, /people/person/nationality, United States)} essentially because the embedding representations of sparse entities are not informative. Concretely, RotatE gives a filtered rank of 1154. Compare this to our method which utilizes the rule that says musicians have the same nationality as the country of the town they are originally from. Having $V_2$ as \texttt{New Orleans} satisfies the rule for the test triplet and our method improves the subject prediction rank to 68, a gain of 1086. An observation about the tail prediction task, i.e. \texttt{(Louis Armstrong, /people/person/nationality, ?)} conveys that baseline methods perform well when entities are not so sparse for e.g. here it is $\texttt{United States}$ that is not sparse. The other examples show a similar qualitative trend demonstrating that using feedback from relevant rules improves embedding representations of entities especially sparse ones.

% % Please add the following required packages to your document preamble:
% % \usepackage{booktabs}
% % \usepackage{graphicx}
% \begin{table*}[h]
% \centering
% \renewcommand{\arraystretch}{1.2}
% \caption{}
% \resizebox{\textwidth}{!}{%
% \begin{tabular}{@{}clcccc@{}}
% \toprule
%  &
%   Rule &
%   \begin{tabular}[c]{@{}c@{}}Rule \\ Support\end{tabular} &
%   \begin{tabular}[c]{@{}c@{}}No.\\ Predictions\end{tabular} &
%   \begin{tabular}[c]{@{}c@{}}Standard \\ Confidence\end{tabular} &
%   \begin{tabular}[c]{@{}c@{}}Embedding \\ Confidence\end{tabular} \\ \midrule
% 1 &
%     /olmpic\_sport./affiliation\_country($V_0, V_2$) $\land$ /location/import\_and\_exports($V_2, V_1$) $\rightarrow$ /olympic\_sport./affiliation\_country($V_0, V_1$)&
%     548 &
%     1723 &
%     0.318 &
%     0.207 \\
% 2 &
%   /olympic\_sport./affiliation\_country($V_0, V_2$) $\land$ /location./adjoins\_location($V_2, V_1$) $\rightarrow$ /olympic\_sport./affiliation\_country($V_0, V_1$)  &
%   1524 &
%   4747 &
%   0.321 &
%   0.404 \\
% % 2 &
% %   /olympic\_games./sport($V_2, V_0$) $\land$ /olympic\_games./participating\_country($V_2, V_1$) $\rightarrow$ /olympic\_sport./affiliation\_country($V_0, V_1$) &
% %   1246 &
% %   3845 &
% %   0.324 &
% %   0.585 \\

% 3 &
%   /sport./pro\_olympic\_athlete($V_0, V_2$) $\land$ /person./nationality($V_2, V_1$) $\rightarrow$ /olympic\_sport./affiliation\_country($V_0, V_1$)  &
%   588 &
%   1792 &
%   0.328 &
%   0.701 \\ \bottomrule
% \end{tabular}%
% }
% \label{tab:gen_rules}
% \end{table*}

% \usepackage{graphicx}
\begin{table*}[ht]
\centering
\caption{Qualitative results showing the comparison between standard embeddding confidence scores for three candidate rules inferring the same head atom. }
\resizebox{\textwidth}{!}{%
\begin{tabular}{cccccc}
\toprule
 &
  Rule &
  \begin{tabular}[c]{@{}c@{}}Rule \\ Support\end{tabular} &
  \begin{tabular}[c]{@{}c@{}}No.\\  Predictions\end{tabular} &
  \begin{tabular}[c]{@{}c@{}}Standard\\ Confidence\end{tabular} &
  \begin{tabular}[c]{@{}c@{}}Embedding\\ Confidence\end{tabular} \\ \hline
1 &
  \begin{tabular}[c]{@{}c@{}}/olmpic\_sport./affiliation\_country($V_0, V_2$) $\land$ /location/import\_and\_exports($V_2, V_1$) $\rightarrow$ \\ /olympic\_sport./affiliation\_country($V_0, V_1$)\end{tabular} &
  548 &
  1723 &
  0.318 &
  0.207 \\
2 &
  \begin{tabular}[c]{@{}c@{}}/olympic\_sport./affiliation\_country($V_0, V_2$) $\land$ /location./adjoins\_location($V_2, V_1$) $\rightarrow$ \\ /olympic\_sport./affiliation\_country($V_0, V_1$)\end{tabular} &
  1524 &
  4747 &
  0.321 &
  0.404 \\
3 &
  \begin{tabular}[c]{@{}c@{}}/sport./pro\_olympic\_athlete($V_0, V_2$) $\land$ /person./nationality($V_2, V_1$) $\rightarrow$ \\ /olympic\_sport./affiliation\_country($V_0, V_1$)\end{tabular} &
  588 &
  1792 &
  0.328 &
  0.701 \\ \bottomrule
\end{tabular}%
}
\label{tab:gen_rules}

\label{tab:my-table}
\end{table*}
% \usepackage{booktabs}
% \usepackage{multirow}
\begin{table}[ht]
\centering
\caption{TransE with different negative sampling techniques. }
\small
\renewcommand{\arraystretch}{.9} % Default value: 1
\begin{tabular}{lcc|cc}
\toprule
\multirow{2}{*}{\begin{tabular}[c]{@{}l@{}}\end{tabular}} & \multicolumn{2}{c|}{FB15K-237} & \multicolumn{2}{c}{WN18RR} \\  \cmidrule(l){2-5} & MRR & Hit@10 & MRR & Hit@10 \\ \midrule
uniform  & 0.241  & 0.422        & 0.186 & 0.459\\
KBGAN  & 0.278 & 0.453           & 0.210 & 0.479\\
self-adversarial & 0.298 & 0.475   & 0.223 & 0.510\\ \midrule
uniform + our method & 0.435 & 0.545         & 0.234 & 0.515\\
\bottomrule
\end{tabular}
\label{tab:neg_sampling}
\end{table}

Next we show qualitatively how rule quality can be improved with embedding guidance through our embedding confidence measure (Eq. \ref{eq:emb_measure}). Table \ref{tab:gen_rules} shows three different rule bodies mined by AIME+ on FB15K-237, which all result in the same head atom i.e. /olympic\_sport./affiliation\_country($V_0, V_1$) which means $V_0$ is an olympic sport that has representation for country $V_1$. An example triplet in $G_{train}$ is (\texttt{Trialathon, /olympic\_sport/affiliation\_country, France}). Rules one and two say that an olympic sport $V_0$ has representation from country $V_1$ if there is another country $V_2$ that represents $V_0$ and is adjoining $V_1$ or does trade with it. Both rules get very similar standard confidence measures because the ratio of corresponding rule support and number of predictions is alike and choosing a relevant rule is not clear. Compare this with rule three, which says an olympic sport $V_0$ has representation from country $V_1$ if there is a professional olympic athlete $V_2$ who plays $V_0$ and $V_2$ has nationality $V_1$. This rule is more apt compared to the earlier ones but still gets similar standard confidence of 0.328. Now contrast this with the embedding confidence scores for each of the three rules. Rule three gets a much better EC score compared to one and two because most of its instantiations give high embedding scores (see Eq. \ref{eq:emb_measure}). This shows that embedding confidence which is able model the rules probabilistically is helpful when used in conjunction with standard confidence to assess the quality of rules when mining.

\subsection{Connections to Negative Sampling}
The results from the previous section made us look at our paradigm through the lens of negative sampling for explanation. At each iteration, when new inferred triplets are introduced, the underlying uniform negative sampling generates higher quality negatives simply because of the superior positives that are augmented. We further show (Table \ref{tab:neg_sampling}) how our method stacks up against other negative sampling methods like uniform, KBGAN \cite{cai2017kbgan} and self-adversarial \cite{rotate}. For all methods including ours, TransE \cite{transe} is used as the embedding method for a fair comparison. The results indicate that the augmented inferred triplets invariably lead to better quality negatives thus improving the overall train set on which the embeddings are learned. In our opinion, this is also a possible explanation for the significant boost in performance for models like TransE and ComplEx that are actually incapable of modelling symmetry and composition patterns in data \cite{rotate}.

\section{Conclusion}
    In this work, we have developed a hybrid method which utilizes the complementary properties of rules and embeddings. Experimental results empirically support the two main claims we raised (1) structure and richer data quality in training results in superior embedding representations, and (2) incorporation of ``global" KG statistical patterns in rule mining lead to reliable rules. We extensively evaluated our approach with varied experiments and showed its effectiveness. The connections to negative sampling motivate us investigate more deeply about the framework in future for possible theoretical claims and develop other general hybrid models for unifying different learning schemes. 

\section*{Acknowledgment}
This research is supported by NSF under contract numbers CCF-1918483, IIS-1618690, and CCF-0939370. 

% \clearpage
\bibliographystyle{IEEEtran}
\bibliography{icdm}

% \bibitem{b1} G. Eason, B. Noble, and I. N. Sneddon, ``On certain integrals of Lipschitz-Hankel type involving products of Bessel functions,'' Phil. Trans. Roy. Soc. London, vol. A247, pp. 529--551, April 1955.
% \bibitem{b2} J. Clerk Maxwell, A Treatise on Electricity and Magnetism, 3rd ed., vol. 2. Oxford: Clarendon, 1892, pp.68--73.
% \bibitem{b3} I. S. Jacobs and C. P. Bean, ``Fine particles, thin films and exchange anisotropy,'' in Magnetism, vol. III, G. T. Rado and H. Suhl, Eds. New York: Academic, 1963, pp. 271--350.
% \bibitem{b4} K. Elissa, ``Title of paper if known,'' unpublished.
% \bibitem{b5} R. Nicole, ``Title of paper with only first word capitalized,'' J. Name Stand. Abbrev., in press.
% \bibitem{b6} Y. Yorozu, M. Hirano, K. Oka, and Y. Tagawa, ``Electron spectroscopy studies on magneto-optical media and plastic substrate interface,'' IEEE Transl. J. Magn. Japan, vol. 2, pp. 740--741, August 1987 [Digests 9th Annual Conf. Magnetics Japan, p. 301, 1982].
% \bibitem{b7} M. Young, The Technical Writer's Handbook. Mill Valley, CA: University Science, 1989.
% \end{thebibliography}
% \vspace{12pt}
% \color{red}
% IEEE conference templates contain guidance text for composing and formatting conference papers. Please ensure that all template text is removed from your conference paper prior to submission to the conference. Failure to remove the template text from your paper may result in your paper not being published.

\end{document}